%% file: template.tex
\documentclass{article}

\usepackage{arxiv}

\usepackage[utf8]{inputenc} % allow utf-8 input
\usepackage[T1]{fontenc}    % use 8-bit T1 fonts
\usepackage{hyperref}       % hyperlinks
\usepackage{url}            % simple URL typesetting
\usepackage{booktabs}       % professional-quality tables
\usepackage{amsfonts}       % blackboard math symbols
\usepackage{nicefrac}       % compact symbols for 1/2, etc.
\usepackage{microtype}      % microtypography
\usepackage{lipsum}
\usepackage{graphicx}
\graphicspath{ {./images/} }

\usepackage[english]{babel}

\newtheorem{proposition}{Proposition}
 \usepackage{amsmath}

 \usepackage{booktabs}
\usepackage{natbib}
\usepackage{floatrow}
\usepackage{multirow}
\floatstyle{plaintop}
\restylefloat{table}
\usepackage{amsmath}
\DeclareMathOperator*{\argmax}{arg\,max}
\DeclareMathOperator*{\argmin}{arg\,min}
\usepackage{amsfonts}       % blackboard math symbols
\usepackage{bbold}          % 1 symbol
\usepackage{hyperref}       % hyperlinks

\newtheorem{axi}{Axiom}
\usepackage{graphicx,import}

\title{Distilling Influences to Mitigate Prediction Churn in Graph Neural Networks}

\author{
 Andreas Roth \\
  TU Dortmund University\\
  Dortmund, Germany \\
  \texttt{andreas.roth@tu-dortmund.de} \\
   \And
 Thomas Liebig \\
  TU Dortmund University\\
  Lamarr Institute for Machine Learning and Artificial Intelligence\\
  Dortmund, Germany \\
  \texttt{thomas.liebig@tu-dortmund.de} \\
}

\begin{document}
\maketitle
\begin{abstract}
Models with similar performances exhibit significant disagreement in the predictions of individual samples, referred to as prediction churn. Our work explores this phenomenon in graph neural networks by investigating differences between models differing only in their initializations in their utilized features for predictions. We propose a novel metric called Influence Difference (ID) to quantify the variation in reasons used by nodes across models by comparing their influence distribution. Additionally, we consider the differences between nodes with a stable and an unstable prediction, positing that both equally utilize different reasons and thus provide a meaningful gradient signal to closely match two models even when the predictions for nodes are similar. Based on our analysis, we propose to minimize this ID in Knowledge Distillation, a domain where a new model should closely match an established one. As an efficient approximation, we introduce DropDistillation (DD) that matches the output for a graph perturbed by edge deletions. Our empirical evaluation of six benchmark datasets for node classification validates the differences in utilized features. DD outperforms previous methods regarding prediction stability and overall performance in all considered Knowledge Distillation experiments.  
\end{abstract}

% keywords can be removed

\input{chapters/introduction}
\input{chapters/preliminaries}

\input{chapters/analysis}
\input{chapters/drop_distillation}
\input{chapters/experiments}
\input{chapters/conclusion}

\section*{Acknowledgments}
This research has been funded by the Federal Ministry of Education and Research of Germany under grant no. 01IS22094E WEST-AI and in the course of the 6GEM research hub under grant number 16KISK038.

\bibliographystyle{abbrvnat}  

\bibliography{references}

\appendix
\input{chapters/appendix}

\end{document}

%% file: chapters/introduction.tex
\section{Introduction}
Neural networks have achieved remarkable success across various domains~\citep{vaswani2017attention,jumper2021highly,roth2022forecasting}, but their predictions often lack reliability and satisfactory explanations, leading to low trust~\citep{samek2021explaining}.
% Churn
One observed issue contributing to this problem is prediction churn, where models with similar performance exhibit significant variability in their predictions~\citep{summers2021nondeterminism,klabunde_prediction_2022}. This churn occurs even among models with identical hyperparameters differing only in their random initializations~\citep{bhojanapalli2021reproducibility,zhuang2022randomness}. When large portions of correctly classified data are misclassified upon model retraining, the models' reliability, trustworthiness, and explainability are reduced.
Churn has been observed across various domains and data structures, but understanding its underlying cause remains challenging. This work focuses on node classification with graph neural networks (GNNs), a domain where churn is particularly prevalent~\citep{schumacher2022effects,klabunde_prediction_2022}.
% Our contributions
Our work investigates the reason behind prediction churn in GNNs by comparing the influence of context nodes on predictions.
We propose a novel metric, the Influence Difference (ID), which allows us to compare the exploited features for a given prediction between a pair of models and empirically verify this prevalence.
Contrary to previous investigations, we hypothesize that stable and unstable predictions exhibit similar differences in their utilized features.
We further hypothesize that stable nodes possess redundant features, allowing their stable prediction even when the features utilized for the prediction change.
To empirically validate our hypotheses, we introduce additional metrics based on ID.

% KD
Knowledge Distillation (KD) is a promising technique to transfer knowledge from a well-performing teacher model to a newly trained student~\citep{buciluǎ2006model,li2014learning,hinton2015distilling} 
%is a crucial direction where prediction churn is particularly undesirable. 
%The student model is trained to mimic the teacher's predictions, 
This enables model compression for a computationally expensive teacher or regular model updates using new data~\citep{gou2021knowledge}. 
Users expect consistent behavior after each update, so closely matching the teacher's predictions is crucial. Previous work formulated the goal of KD as directly minimizing churn~\citep{jiangchurn}.
Based on our findings, we propose to extend KD %to minimize churn directly and 
by also matching the influences of predictions as a regularization.
As the exact formulation is computationally prohibitive, we introduce DropDistillation (DD), an efficient approximation that mimics the influence of adjacent nodes by removing random edges equally from both the teacher and the student model.
% Experiments
Our empirical analysis validates our hypotheses for several benchmark datasets, further motivating the need for transferring the reasons.
Comparing DropDistillation with several state-of-the-art methods, our approach improves prediction churn between teachers and students and overall performance.
We summarize our key contributions:
\begin{itemize}
    \item We investigate the reason behind prediction churn in GNNs by comparing the influence of context nodes on predictions using a novel metric. We also connect differences in prediction stability with the availability of redundant features (Section~\ref{sec:single}).
    \item Based on our findings, we extend knowledge distillation to minimize churn directly and match the features exploited for predictions as a regularization technique and propose an efficient approximation, namely DropDistillation (DD) (Section~\ref{sec:dd}). 
    \item Our empirical evaluation validates our claims and confirms the effectiveness of DropDistillation for Knowledge Distillation for various benchmark datasets (Section~\ref{sec:experiments}).
\end{itemize}

These contributions aim to enhance our understanding of prediction churn in GNNs, and propose a novel direction to address churn in knowledge distillation, resulting in increased reliability and overall performance.

%% file: chapters/preliminaries.tex
\section{Preliminaries}
We start by introducing basic notations and concepts we use throughout this work.
Let $G =(V,E)$ be a graph containing a set of $n$ nodes $V = \{v_1,\dots,v_n\}$ and a set of edges $E$ indicating the connectivity between pairs of nodes. 
We also express the set of edges as an adjacency matrix $\mathbf{A}\in\mathbb{R}^{n\times n}$ of pairwise connections that may additionally weigh nodes differently.
We consider the task of node classification where training data comes either from other graphs (inductive case) or labels available for a subset of the nodes $V_t\subset V$ (transductive case). Our work investigates graph neural networks (GNNs)~\citep{kipf2016semi}, though only a superficial understanding is needed.
GNNs represent a learnable function $f:\mathbb{R}^{n\times d} \times \mathbb{R}^{n\times n}\to\mathbb{R}^{n\times c}$, mapping a graph signal and an adjacency structure to output logits $f(\mathbf{X},\mathbf{A})=\mathbf{C}$ for $c$ classes and each node. 
$\mathbf{Y}\in\mathbb{R}^{n\times c}$ denotes the true label matrix.

\subsection{Prediction Churn}
Prediction churn~\citep{NIPS2016_dc4c44f6,NIPS2016_dc5c768b} describes a phenomenon in which models make different predictions on the same data points.%, also referred to as prediction instability~\citep{klabunde_prediction_2022}.
%Formal definition
Formally, we define churn for any two functions $f, g$ for node classification mapping the $d$-dimensional graph signal $\mathbf{X}\in\mathbb{R}^{n\times d}$ and the adjacency matrix $\mathbf{A}$ to class probabilities for every node. We define the set of unstable nodes by 
\begin{equation}
    \mathcal{U}_{f,g} = \{v_i\in V_{\mathrm{test}} | \argmax_{c\in C} f(\mathbf{X},\mathbf{A})_{ic} \neq \argmax_{c\in C}g(\mathbf{X},\mathbf{A})_{ic}\}\, .
\end{equation}
We further define $\mathbf{s}_{f,g}\in[0,1]^{n}$ to be the binary vector indicating with a one whether each node was predicted stable between $f$ and $g$.
The pairwise churn is then defined as the ratio of unstable nodes over the total number of test nodes, as given by
\begin{equation}
    C(f,g) = \frac{|\mathcal{U}_{f,g}|}{|\mathcal{V}_{\mathrm{test}}|} = \frac{1}{n} \sum_{v_i\in \mathcal{V}_{\mathrm{test}}} \mathbb{1}_{\{v_i \in \mathcal{U}_{f,g}\}}\, .
\end{equation}
Here, $\mathbb{1}$ denotes the indicator function, that is 1 if the condition is satisfied and 0 otherwise.
Churn is undesirable for many reasons, including the reproducibility of scientific results, reliability, and trust in machine learning models~\citep{jiangchurn,liu2022model}. 
Especially when continuously delivering updated models, the experience should be stable and consistent~\citep{NIPS2016_dc4c44f6}. 

Churn occurs in various scenarios, such as using different model architectures or hyperparameters. But even when these are the same, churn occurs even when the initial parameters are slightly altered~\citep{bhojanapalli2021reproducibility,zhuang2022randomness}. Non-determinism of GPU operations also produces churn even when all initial parameters are the same~\citep{summers2021nondeterminism}.
%Various investigations demonstrated the existence of large churn mainly for grid-structured data~\citep{bhojanapalli2021reproducibility,summers2021nondeterminism}, even when the corresponding accuracy of the classifiers differs marginally.
For graph-structured data, several studies observed an instability of node embeddings that holds for stochastic operations like dropout~\citep{wang2020towards,klabunde_prediction_2022}.
However, the reason behind churn and the inability of models optimized on the same data to develop similar decision rules remains unclear.

\subsection{Knowledge Distillation}
Knowledge Distillation (KD)~\citep{buciluǎ2006model,li2014learning,hinton2015distilling} is one particular domain of interest for reducing churn. Here, the goal is to distill the knowledge of a pre-trained large model, called the teacher $T$, into a smaller model, called the student $S$. In our case, both $S$ and $T$ are functions that perform node classification. 
This is typically used to compress the knowledge for resource efficiency during inference~\citep{cheng2018model} but also to optimize iteratively updated models for continuous deployment~\citep{jiangchurn}.
In general, a distillation loss $\mathcal{L}_{\mathrm{distill}}(S,T)$ is used to match the output or intermediate representations of $T$ and $S$~\citep{li2014learning,hinton2015distilling} in addition to the original classification loss.
Prediction churn and Knowledge Distillation are inherently connected, as they share a common goal. Pairs of models should produce similar predictions, so in a perfect scenario, the churn between student and teacher would be zero. 
%In any case, the underlying goal of Knowledge Distillation is to closely match the outputs of $S$ to a fixed $T$.
%Specifically, the popular Knowledge Distillation uses the logits of the teacher model scaled by a temperature $\tau$ as soft targets using the cross-entropy loss $\mathcal{L}_{\mathrm{CE}}$ in a convex combination with the original loss using a weighting parameter $\alpha\in[0,1]$, resulting in
%\begin{equation}
%\label{eq:kd}
%    \mathcal{L}_{\mathrm{KD}}(S,T) = (1-\alpha) \mathcal{L}_{CE}(\frac{S(\mathbf{X},\mathbf{A})}{\tau},\frac{T(\mathbf{X},\mathbf{A})}{\tau}) + \alpha \mathcal{L}_{\mathrm{CE}}(S(\mathbf{X},\mathbf{A}),\mathbf{Y})\, .
    %= -\frac{1}{N}\sum_{n=1}^N\sum_{c=1}^C \frac{e^{\frac{T(\mathbf{X},G)_{nc}}{\tau}}}{\sum e^{\frac{T(\mathbf{X},G)_{nc}}{\tau}}}\log \frac{e^{S(\mathbf{X},G)/\tau}}{\sum e^{S(\mathbf{X},G)/\tau}}
%\end{equation}

\subsection{Related Work}
Few investigations connected Knowledge Distillation and prediction churn.
\citet{jiangchurn} continuously optimize new models and introduce a churn constraint between consecutive instantiations. They showed that matching the outputs is equivalent to directly minimizing the prediction churn between $S$ and $T$, though they require strong assumptions on the generalization bounds.
\citet{bhojanapalli2021reproducibility} propose a co-distillation approach that optimizes two models simultaneously while matching their outputs. 
Other methods similarly propose to reduce churn solely based on the outputs~\citep{NIPS2016_dc5c768b,summers2021nondeterminism}. 
Our investigation also builds on ideas from quantifying the similarity of neural networks~\citep{lenc2015understanding,klabunde2023similarity}. In particular, \citet{jones2022if} determine the similarity between two models using vectorized saliency maps. \citet{allen2020towards} propose a theory in which models would learn different subsets of the available features but not all of them, as they are not required to correctly classify a large part of the data.
We extend these approaches to the specific properties of graph-structured data and relate them to prediction churn.
%As we show next, the connection between churn and single-view data is more central, with churn only being a consequence.  
%methods to determine churn
%This will allow us to design novel approaches to reduce it more effectively and apply this to align the student model to its teacher in Knowledge Distillation.

%% file: chapters/analysis.tex
\section{Comparing Differences in Influence of Predictions}
\label{sec:single}

\begin{figure}
    \centering
    \def\svgwidth{\columnwidth}
    \resizebox{\textwidth}{!}{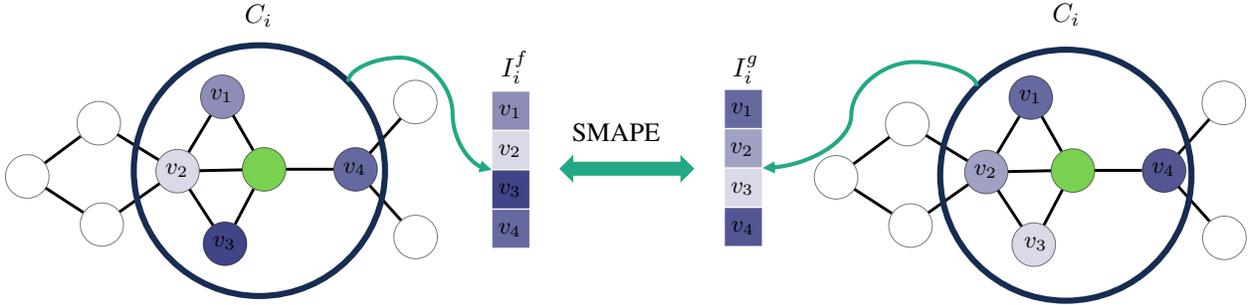}
    
    \caption{Influence Difference for the green root node between two models $f$ (left) and $g$ (right). Context nodes $C_i$ are encircled, and the different shades of blue represent its nodes' value in the influence distribution. Symmetric Mean Absolute Percentage Error (SMAPE) between the two influence vectors is calculated.}
    \label{fig:enter-label}
\end{figure}
%We sometimes refer to this influence as the reason behind a prediction
We now delve into the underlying reason behind prediction churn in the context of node classification with graph neural networks. 
Given a single node, we want to investigate what makes its prediction stable or unstable across models.
We build our investigations on recent progress in determining the similarity between neural networks~\citep{allen2020towards}.
It was found that comparing the outputs or representations of neural networks alone overestimates their similarity due to the correlation between data points~\citep{jones2022if}. In our case, churn may be insufficient in determining the actual similarity between pairs of models. 
A recent theory indicates that models may only learn a subset of the available features from the data depending on their parameter initialization. %However, empirical confirmation and quantifying metrics still need to be improved. 
Confirming these ideas would provide insights into the causes of prediction churn. %, as only these nodes having access to the utilized features would be classified correctly.
%Considering our task of node classification, 
We incorporate these ideas and propose the following statement that we will empirically investigate later:

\begin{axi}[A1]
\label{hyp:diff}
    Prediction churn is a consequence of models utilizing different features for their predictions.
\end{axi}

Thus, even when two models have low prediction churn, they may contain different knowledge.
Depending on which features a model utilizes, the predicted class may change. Consequently, our objective is to quantify the difference between features used by nodes to make specific predictions between a pair of models. 
We also aim to investigate the distinguishing factors between stable and unstable nodes in terms of their utilized features.
Analyzing this behavior would allow us to construct better-informed methods to mitigate the resulting churn and improve the overall reliability of node classification, e.g., in Knowledge Distillation.
Building on these insights, we now propose metrics to quantify the difference in reasoning between pairs of models for their node predictions.

%

%the importance of not only considering the outputs when determining the similarity between neural networks but to include the gradients in some way.

%We now investigate the underlying reason behind prediction churn.
%On the one hand, we have the theoretical insights that models only learn a subset of the available features that are needed to classify a large part of the data correctly.
%On the other hand, we have the empirical observation that models produce unstable predictions.

%The theoretical investigation~\cite{allen2020towards} demonstrated that learned feature subsets differ for models with varying random initializations, resulting in the misclassified elements differing based on their available subset of views.%, referred to as single-view data. 
%Thus, these single-view nodes are a reason for model churn in GNNs.
%This poses a risk to test data as they may not be classified consistently and reliably. 

%Therefore, churn would also be reduced when two models learn the same views from the data. However, the contrary is not true, as models can predict the same outcome while relying on different but redundant views.
%We argue that churn alone is insufficient to capture the underlying differences between pairs of models.

\subsection{Comparing the Reasons behind Predictions}
Instead of comparing the output differences for a pair of models, we propose to compare which features each model utilized for its predictions directly.
Our data's graph structure and the node classification task pose a challenge.
Given the prediction for a root node, we want to determine its influencing features and compare this between models. As all neighboring nodes can affect the prediction to some degree, we propose to view each neighboring node as one potential feature the root node can utilize. 
When a root node relies on different neighbors for distinct models, this indicates that models learn disjoint feature subsets and would thus be more meaningful than churn $C$.  

%As we consider the task of node classification in this work, we compare the importance of adjacent nodes on the output of a given root node. 
%Toward this, we next introduce two new metrics quantifying this idea.

Formally, we want to determine the importance of an initial node signal $\mathbf{x}_j$ at node $v_j$ on the extracted representation $\mathbf{z}_{i}$ of node $v_i$.
As the gradient $\frac{\partial z_{ia}}{\partial x_{jb}}$ represents the sensitivity of the a-th feature of node $i$'s representations $z_{ia}$ to a variable $x_{jb}$, our requirements are met by the well-established influence scores~\citep{xu2018representation}
\begin{equation}
    I(i,j) = \sum_{a=1}^{d^\prime} \sum_{b=1}^{d} \left|\frac{\partial z_{ia}}{\partial x_{jb}}\right|\, .
\end{equation}
for node $v_j$ on node $v_i$.
These sum the gradient magnitudes of each logit output $z_{ia}$ to each input feature $x_{jb}$.
%Influence Distribution
Following \citet{xu2018representation}, the influence distribution is then defined as the normalized influences 
\begin{equation}
    I_i(j) = \frac{I(i,j)}{\sum_{k\in C_i} I(i,k)}
\end{equation}
over all context-nodes $C_i$ of a root node $v_i$. We further denote the influence distribution of a given model $f$ as $I^f_i(j)$. We now propose two novel metrics that compare the differences between models based on the influence distribution.

%\subsubsection{Influence Difference}
We propose a generic metric that compares the deviations between each value in the influence distributions $I^f_i(j)$ of model $f$ and $I^g_i(j)$ of model $g$. Since the magnitude of each normalized score $I_i(j)$ depends on the number of context nodes $|C_i|$, we use a relative metric.
In general, $f$ and $g$ are commutable, so the metric should also be symmetric.
Thus, we use the symmetric mean absolute percentage error (SMAPE)~\citep{chenandyang}, which has a direct interpretation. We define the expected Influence Difference as
\begin{equation}
    \mathrm{ID}(f,g) = \mathbb{E}_{v_i \sim V,j\sim C_i}\left[\frac{|I^f_i(j) - I^g_i(j)|}{0.5\cdot(|I^f_i(j)| + |I^g_i(j)|)}\right] %= \frac{1}{|V|}\sum_{i\in V}\frac{1}{|C_i|}\sum_{j \in C_i} \frac{|I^f_i(j) - I^g_i(j)|}{0.5\cdot(|I^f_i(j)| + |I^g_i(j)|)} 
\end{equation}
between a pair of functions $f$ and $g$ with shared domain and codomain. 
This metric should be small when both $f$ and $g$ rely on the same nodes and large when they rely on different nodes for their representations. Figure~\ref{fig:enter-label} provides a visualization of this metric for one root node.
A key property of ID is that even when the prediction churn is small, it can still provide meaningful information about the differences between the knowledge acquired between the two models:
\begin{proposition}
\label{prop:imply}
    Given $f$, $g$. $C(f,g)=0$ does not imply a low Influence Difference.
\end{proposition} 

Given these properties, we think ID provides more profound insights about which features models rely on and how stable that is between models.

%\subsubsection{Influence Churn}
%While the magnitude of the influence distributions may change, the ordering of importance may still be preserved.
%Thus, we also want to understand which nodes are the most relevant for the output of a node and whether these differ between initializations. Similar to model churn, we introduce the influence churn
%\begin{equation}
%    \mathrm{IC}(f,g) = \mathbb{E}_{v_i \sim V}\left[\argmax_{j \in C_i} I_i^{f}(j) \neq \argmax_{j \in C_i}I_i^{g}(j)\right]
%\end{equation}
%as the disagreement of the most influential other context node $j\in C_i$ for a node root node among models $f$ and $g$. The $IC$ is small if all features or the same features in the data are utilized.

\subsection{Differences between Stable and Unstable Nodes}
\label{sec:inf_cond}

Our proposed metric, ID, allows us to determine the difference in influence globally over all nodes between a pair of models. 
Previous work based on the churn metric showed that only some nodes are predicted unstable, while others are predicted stable across many models~\citep{klabunde_prediction_2022}.
However, if models learn different feature subsets, 
this would hold for all nodes, not only those with an unstable prediction. Thus, we make the following claim:

%We additionally want to investigate whether this also holds for influence on a prediction. If the theory regarding multi-view data~\cite{allen2020towards} holds, all elements would similarly rely on different subsets of the available views.
%different subsets of features

\begin{axi}[A2]
\label{hyp:unconditional}
    Stable and unstable nodes have a similar Influence Difference.
\end{axi}

To verify this statement empirically, we calculate the correlation between the average influence differences $\mathbf{id}\in\mathbb{R}^n$ for each node and the stability $\mathbf{s}\in\{0,1\}^{n}$ of each node's prediction. We utilize Pearson's correlation coefficient
\begin{equation}
    \mathrm{corr}(\mathbf{id}, \mathbf{s}) = \frac{\mathrm{cov}(\mathbf{id},\mathbf{s})}{\sigma_\mathbf{id} \sigma_\mathbf{s}}
\end{equation}
based on the covariance cov, and the standard deviations $\sigma_{\mathbf{id}}$ and $\sigma_\mathbf{s}$. A high correlation would indicate that stable nodes also have larger differences in influence between models. Thus, we expect them to be uncorrelated.

%we calculate $ID$ separately for the set of nodes $U(f,g)$ with an unstable prediction and the complement set of stable nodes $V\setminus U(f,g)$.
%Formally, we condition the expected Influence Difference on the unstable nodes 
%\begin{equation}
%    ID^U(f,g) = \mathbb{E}_{v_i \sim V,j\sim C_i}\left[\frac{|I^f_i(j) - I^g_i(j)|}{0.5\cdot(|I^f_i(j)| + |I^g_i(j)|)}\middle\vert v_i \in U(f,g)\right]\, .
%\end{equation}
%Similarly, we define the influence difference conditioned $ID^S(f,g)$ on the stable nodes.
    
%We also define the influence churn conditioned on unstable nodes
%\begin{equation}
%    \mathrm{IC}^U(f,g) = \mathbb{E}_{v_i \sim V}\left[\argmax_{j \in C_i} I_i^{f}(j) \neq \argmax_{j \in C_i}I_i^{g}(j) \middle\vert v_i \in U(f,g)\right]
%\end{equation}
%and similarly for stable nodes $IC^S(f,g)$.

\subsection{Feature Redundancy Stabilizes Predictions}

The question remains what leads to the unstable prediction of some of the nodes. As outlined by \citet{allen2020towards} regarding model similarity, some elements may contain redundant features.
For node classification, our interpretation is that stable nodes similarly have access to redundant features. Thus a similar Influence Difference has a smaller effect on the actual prediction. We propose the following statement:

\begin{axi}[A3]
\label{hyp:redundant}
    Stable nodes have access to more redundant features.
\end{axi}

%Another aspect we would like to understand is if unstable nodes actually correspond to single-view nodes and have fewer views available.
For verification, we need to determine the number of redundant features each node has available for their predictions. However, the discriminating features are not observable in the data, so we use an indicator that should closely correlate with the number of available features.
We utilize the number of context nodes with the same label, as these should provide redundant signals.
We let $D_i(y)$ be the ratio of each label $y\in c$ in the context $C_i$ of node $v_i$.
We calculate the entropy
\begin{equation}
H(i) = - \sum_{y \in c} \log(D_i(y)) D_i(y)
\end{equation}
of these label ratios for each node $i$. 
Low entropy corresponds to redundant views as the prediction could rely on features from different nodes. We denote the vector of label entropies for all nodes as $\mathbf{h} = (H(1),\dots,H(n))^T$.
To validate A\ref{hyp:redundant}, we again calculate the correlation to the node stability vector $\mathbf{s}$ using Pearson's correlation coefficient
\begin{equation}
    \mathrm{corr}(\mathbf{h}, \mathbf{s}) = \frac{\mathrm{cov}(\mathbf{h},\mathbf{s})}{\sigma_\mathbf{h} \sigma_\mathbf{s}}
\end{equation}
based on their covariance $\mathrm{cov}$ and respective standard deviations $\sigma_\mathbf{h}$ and $\sigma_\mathbf{s}$.
A high correlation indicates stable nodes would have more variance in their neighboring labels. Thus, if A\ref{hyp:redundant} holds, we expect an anti-correlation.

%Again, we condition the label entropy on the stable nodes
%\begin{equation}
%    H^S = E_{v_i\sim V}\left[H(i)\middle\vert v_i \notin U(f,g)\right]
%\end{equation}
%and compare this with the entropy conditioned on the unstable nodes
%\begin{equation}
%H^U = E_{v_i\sim V}\left[H(i)\middle\vert v_i \in U(f,g)\right]\, .
%\end{equation}

%% file: chapters/influence_difference.pdf_tex
%% Creator: Inkscape 1.2.2 (732a01da63, 2022-12-09), www.inkscape.org
%% Accompanies image file 'influence_difference.pdf' (pdf, eps, ps)
\begingroup%
  \makeatletter%
  \providecommand\color[2][]{%
    \errmessage{(Inkscape) Color is used for the text in Inkscape, but the package 'color.sty' is not loaded}%
    \renewcommand\color[2][]{}%
  }%
  \providecommand\transparent[1]{%
    \errmessage{(Inkscape) Transparency is used (non-zero) for the text in Inkscape, but the package 'transparent.sty' is not loaded}%
    \renewcommand\transparent[1]{}%
  }%
  \providecommand\rotatebox[2]{#2}%
  \newcommand*\fsize{\dimexpr\f@size pt\relax}%
  \newcommand*\lineheight[1]{\fontsize{\fsize}{#1\fsize}\selectfont}%
  \ifx\svgwidth\undefined%
    \setlength{\unitlength}{3165.75bp}%
    \ifx\svgscale\undefined%
      \relax%
    \else%
      \setlength{\unitlength}{\unitlength * \real{\svgscale}}%
    \fi%
  \else%
    \setlength{\unitlength}{\svgwidth}%
  \fi%
  \global\let\svgwidth\undefined%
  \global\let\svgscale\undefined%
  \makeatother%
  \begin{picture}(1,0.25136224)%
    \lineheight{1}%
    \setlength\tabcolsep{0pt}%
    \put(0,0){\includegraphics[width=\unitlength,page=1]{influence_difference.pdf}}%
    \put(0.272362,0.10589908){\makebox(0,0)[lt]{\lineheight{1.25}\smash{\begin{tabular}[t]{l}\footnotesize$v_4$\end{tabular}}}}%
    \put(0,0){\includegraphics[width=\unitlength,page=2]{influence_difference.pdf}}%
    \put(0.16508742,0.16465293){\makebox(0,0)[lt]{\lineheight{1.25}\smash{\begin{tabular}[t]{l}\footnotesize$v_1$\end{tabular}}}}%
    \put(0,0){\includegraphics[width=\unitlength,page=3]{influence_difference.pdf}}%
    \put(0.12909287,0.10447761){\makebox(0,0)[lt]{\lineheight{1.25}\smash{\begin{tabular}[t]{l}\footnotesize$v_2$\end{tabular}}}}%
    \put(0,0){\includegraphics[width=\unitlength,page=4]{influence_difference.pdf}}%
    \put(0.16735868,0.04596067){\makebox(0,0)[lt]{\lineheight{1.25}\smash{\begin{tabular}[t]{l}\footnotesize$v_3$\end{tabular}}}}%
    \put(0,0){\includegraphics[width=\unitlength,page=5]{influence_difference.pdf}}%
    \put(0.92348259,0.10518834){\makebox(0,0)[lt]{\lineheight{1.25}\smash{\begin{tabular}[t]{l}\footnotesize$v_4$\end{tabular}}}}%
    \put(0,0){\includegraphics[width=\unitlength,page=6]{influence_difference.pdf}}%
    \put(0.8162497,0.16370528){\makebox(0,0)[lt]{\lineheight{1.25}\smash{\begin{tabular}[t]{l}\footnotesize$v_1$\end{tabular}}}}%
    \put(0,0){\includegraphics[width=\unitlength,page=7]{influence_difference.pdf}}%
    \put(0.78027008,0.10376688){\makebox(0,0)[lt]{\lineheight{1.25}\smash{\begin{tabular}[t]{l}\footnotesize$v_2$\end{tabular}}}}%
    \put(0,0){\includegraphics[width=\unitlength,page=8]{influence_difference.pdf}}%
    \put(0.397638,0.15185975){\makebox(0,0)[lt]{\lineheight{1.25}\smash{\begin{tabular}[t]{l}\footnotesize$v_1$\end{tabular}}}}%
    \put(0.39716418,0.12011372){\makebox(0,0)[lt]{\lineheight{1.25}\smash{\begin{tabular}[t]{l}\footnotesize$v_2$\end{tabular}}}}%
    \put(0.39716418,0.08836769){\makebox(0,0)[lt]{\lineheight{1.25}\smash{\begin{tabular}[t]{l}\footnotesize$v_3$\end{tabular}}}}%
    \put(0.39716418,0.05662165){\makebox(0,0)[lt]{\lineheight{1.25}\smash{\begin{tabular}[t]{l}\footnotesize$v_4$\end{tabular}}}}%
    \put(0,0){\includegraphics[width=\unitlength,page=9]{influence_difference.pdf}}%
    \put(0.58440654,0.15328121){\makebox(0,0)[lt]{\lineheight{1.25}\smash{\begin{tabular}[t]{l}\footnotesize$v_1$\end{tabular}}}}%
    \put(0.58393272,0.12153518){\makebox(0,0)[lt]{\lineheight{1.25}\smash{\begin{tabular}[t]{l}\footnotesize$v_2$\end{tabular}}}}%
    \put(0.58440654,0.08978915){\makebox(0,0)[lt]{\lineheight{1.25}\smash{\begin{tabular}[t]{l}\footnotesize$v_3$\end{tabular}}}}%
    \put(0.58488036,0.05804312){\makebox(0,0)[lt]{\lineheight{1.25}\smash{\begin{tabular}[t]{l}\footnotesize$v_4$\end{tabular}}}}%
    \put(0,0){\includegraphics[width=\unitlength,page=10]{influence_difference.pdf}}%
    \put(0.19308458,0.2262497){\makebox(0,0)[lt]{\lineheight{1.25}\smash{\begin{tabular}[t]{l}$C_i$\end{tabular}}}}%
    \put(0.84325987,0.22696044){\makebox(0,0)[lt]{\lineheight{1.25}\smash{\begin{tabular}[t]{l}$C_i$\end{tabular}}}}%
    \put(0.39939824,0.18336887){\makebox(0,0)[lt]{\lineheight{1.25}\smash{\begin{tabular}[t]{l}$I_i^f$\end{tabular}}}}%
    \put(0.58596069,0.18336887){\makebox(0,0)[lt]{\lineheight{1.25}\smash{\begin{tabular}[t]{l}$I_i^g$\end{tabular}}}}%
    \put(0.45661457,0.13124852){\makebox(0,0)[lt]{\lineheight{1.25}\smash{\begin{tabular}[t]{l}SMAPE\end{tabular}}}}%
    \put(0.82071905,0.04517513){\makebox(0,0)[lt]{\lineheight{1.25}\smash{\begin{tabular}[t]{l}\footnotesize$v_3$\end{tabular}}}}%
  \end{picture}%
\endgroup%

%% file: chapters/drop_distillation.tex
\section{Aligning the Influence in Knowledge Distillation}
\label{sec:dd}
%While reliable predictions are generally desired, w
We now build on our claims to mitigate churn in Knowledge Distillation (KD). Here, we assume the teacher model $T$ is given and exhibits desirable performance and reasonings for its predictions.
Current state-of-the-art for reducing prediction churn in KD only matches the outputs or representations, e.g., by directly minimizing prediction churn~\citep{jiangchurn}. 
Based on our claimed Axioms, this may not transfer the knowledge in $T$ to $S$ as different features can be exploited for the same predictions, hurting generalizability (see Proposition~\ref{prop:imply}). 
%While other approaches propose to additionally match intermediate representations~\citep{joshi2022representation}, our investigation holds as multiple differing feature combinations can similarly result in the same representations.

%Also based on our hypotheses.
We propose minimizing ID to directly match the reasonings behind predictions and an improved knowledge transfer
If we could achieve the student to mimic the reasoning of a high-capacity teacher, the performance, reliability, and generalization should be improved.
Critically, based on Axiom~\ref{hyp:unconditional}, nodes with the same prediction still provide a strong gradient signal when optimizing for ID. Here, optimizing churn would provide a negligible gradient signal.
Consequently, optimizing ID allows for transferring more of the utilized features from the teacher.
%or intermediate representations but do not consider matching the influence. Matching the outputs also suffers over-confidence, which occurs when the teacher is certain in its predictions, and its output does not contain additional signals to the true labels~\cite{hinton2015distilling}.
%Critically, the empirical results indicate that minimizing the $ID$ for multi-view nodes also improves the $ID$ for single-view nodes. Contrarily, when directly minimizing $C$, the churn for multi-view nodes becomes close to zero without capturing all views.
%Based on these findings, we now propose a new method that minimizes the difference in influence distributions to aid Knowledge Distillation.   
%We assume the teacher model $T$ to exhibit better reasoning than a student $S$, i.e., the teacher is an ensemble or a more expressive model.
Formally, we extend any given distillation loss $\mathcal{L}_{\mathrm{distill}}$ matching representations by our ID term matching influences, resulting in the regularized optimization problem
\begin{equation}
    S^* = \argmin_{S\in\mathcal{S}} \mathcal{L}_{\mathrm{distill}}(S,T) + \mathrm{ID}(S,T)\, .
    \label{eq:min_id}
\end{equation}

We note that matching the influence distributions is related to matching Jacobians, which has been explored for grid-structured data~\citep{czarnecki2017sobolev,zagoruyko2017paying,srinivas2018knowledge}. 
%However, these approaches are unfit for graph neural networks.
The unique properties of the node classification task make it prohibitive to calculate $\mathrm{ID}(S,T)$ exactly as the number of gradients needed is $(n\times d)\cdot (n \times h)$ when considering all pairs of node representations $\mathbf{H}^l\in\mathbb{R}^{n\times h}$ and input features $\mathbf{X}\in\mathbb{R}^{n\times d}$. 
Instead, we propose to approximate ID, which we will describe in detail next. 

\subsection{Approximating the Influence Difference using DropDistillation}
Existing work on Knowledge Distillation for image classification proposed approximating the Jacobians by applying Gaussian noise to the input features~\citep{srinivas2018knowledge,nam2021diversity}. %Their motivation is to not only match the outputs for the true data points but also match closeby points. 
However, given the smoothing properties of graph neural networks, known as over-smoothing~\citep{oono2019graph,roth2022transforming,roth2023rank}, these high-frequency signals are unfit for GNNs as they are filtered out quickly and have little effect on the output. This was similarly argued by \citet{nam2021diversity} for grid-structured data. 

Instead, we argue that a broader class of perturbations can be used to approximate the gradients: When the influence distributions of the teacher and the student are close, any perturbations of the input should lead to a similar output. Input perturbations were shown to be beneficial for learning across multiple domains~\citep{Rong2020DropEdge,roth2021data}. 
Our approach uses a perturbation that exploits some of the properties specific to graphs by removing edges of the underlying graph, which we call DropDistillation (DD). 
Our intuition matches our goal: When removing an edge, the influence of the adjacent node is reduced, while the influence of all other edges should increase correspondingly. When repeatedly removing different edges and matching the outputs, the student can learn to mimic the influence of particular neighbors for the teachers' prediction of a node.
%Other graph augmentations, e.g., node dropout or feature masking~\cite{you2020graph}, could be similarly used to approximate matching the influence. We describe the details of our method next.

\subsection{DropDistillation} 
For each training step, we remove edges uniformly at random with probability $p^*$. We define the set of edges to drop $\mathcal{E}^{drop} = \{e_{uv}\in\mathcal{E} | p_{uv} < p^*\}$ using uniformly random values $p_{uv}\sim U(0,1)$ for each edge $(u,v)\in \mathcal{E}$.
We note that dropping edges were already successfully employed in DropEdge~\citep{Rong2020DropEdge}, though its reason is fundamentally different, as they are trying to mitigate over-smoothing and over-fitting.
While their approach aims to map all perturbed graphs to the same output, we explicitly want different outputs to match the influence distribution.
We express the perturbation using a matrix $\mathbf{A}^{\mathrm{drop}}$ that offsets the edges to drop, i.e. $\mathbf{A}^{drop}_{uv} = -\mathbf{A}_{uv}$ for all $(u,v)\in \mathcal{E}^{\mathrm{drop}}$. All other edges are scaled up accordingly, e.g., by recalculating the remaining edges' mean or attentional coefficients. In this case, the expected edge strength remains the same, as was demonstrated in DropEdge~\citep{Rong2020DropEdge}.

Our DropDistillation (DD) matches the logit outputs of $T$ and $S$ for the perturbed inputs using the mean squared error 
\begin{equation}
    %ID(S,T) \propto 
    \mathcal{L}_{DD}(S,T) = \frac{1}{N \cdot C}\sum_{v,i=1}^{N,C} \left(S(\mathbf{X},\mathbf{A}+\mathbf{A}^{\mathrm{drop}})_{vi} - T(\mathbf{X}, \mathbf{A}+\mathbf{A}^{\mathrm{drop}})_{vi}\right)^2
\end{equation}
across all $n$ nodes and $c$ classes.
We use $\mathcal{L}_{DD}$ in combination with a given method for Knowledge Distillation as described in Eq.~\ref{eq:min_id}. 
Since we use an approximation of the influence distributions, the resulting inputs may not be representative of the distribution of the actual gradients, and minimizing $\mathcal{L}_{DD}(S,T)$ may not lead to optimal results. 
To mitigate this, we start by solely optimizing $\mathcal{L}_{DD}$ until the loss plateaus and most of the teacher's reasoning is transferred to the student. As the student is typically unable to match the reasoning completely, we fine-tune the student using the regular distillation loss $\mathcal{L}_{\mathrm{distill}}$.% to achieve a locally optimal distillation loss. 
%This also matches the theoretical insights on multi-view data well. 
%Optimizing the classification loss would only lead to a subset of the features to be learned~\cite{allen2020towards}. However, by matching the influence of the teacher, its potentially more extensive set of features can be learned by the student. 
This also has a positive effect on runtime, as only a single forward pass is needed in each step.
%We next present a theoretical motivation for our approximation.

\subsection{The Theoretical Motivation for DropDistillation}
We provided some intuition for dropping edges to approximate the influence distribution, for which we now also want to give a theoretical reason. This also motivates our usage of the squared error function.
\citet{srinivas2018knowledge} show that adding random noise can approximately match the Jacobians using the first-order Taylor approximation. This was also adapted by \citet{nam2021diversity}. 
We formally show that dropping edges similarly match the gradients and thus the influence difference.
%While the Taylor approximation may not be close in our case, it provides some insights.  
The critical similarity is that for many aggregation functions, e.g., mean or using attention, we can still assume zero expectation for edge perturbations $\mathbb{E}(\mathbf{A}_{\mathrm{drop}})=\mathbf{0}$, as the remaining edges are scaled up equally.
% Thus, our method allows the proof for random noise~\citep{srinivas2018knowledge} with slight modifications for our proposed method:
Our proof closely follows the proof for random noise~\citep{srinivas2018knowledge}:

\begin{proposition} Let $T,S:\mathbb{R}^{n\times d} \times \mathbb{R}^{n\times n}\to\mathbb{R}^{n\times c}$ be functions, and let $\mathbf{X}\in\mathbb{R}^{n \times d},\mathbf{A}\in\mathbb{R}^{n\times n},\mathbf{A}_{\mathrm{drop}}\in\mathbb{R}^{n \times n}$ be matrices. We further assume $\mathbb{E}(\mathbf{A}^{\mathrm{drop}}_{vu})=0$ for all $v,u\in[1,\dots N]$. Then,
\begin{align*}
    &\mathbb{E}\left[\sum_{v,i=1}^{n,c}(T(\mathbf{X}, \mathbf{A}+\mathbf{A}_{\mathrm{drop}})_{vi} - S(\mathbf{X},\mathbf{A}+\mathbf{A}_{\mathrm{drop}})_{vi})^2\right]  = \sum_{v,i=1}^{n,c}(T(\mathbf{X})_{vi} - S(\mathbf{X})_{vi})^2 \\
    &+ \mathbb{E}_{\mathbf{A}_\mathrm{drop}}\left[ \left(vec(\nabla_{\mathbf{A}} T(\mathbf{X},\mathbf{A})_{vi} - \nabla_{\mathbf{A}} S(\mathbf{X},\mathbf{A})_{vi})^T vec(\mathbf{A}_{\mathrm{drop}})\right)^2 + \sum_{v,u=1}^{n,n}\mathcal{O}((\mathbf{A}_{\mathrm{drop}})_{vu}^2)\right]
\end{align*}
\end{proposition}
We provide the detailed proof as supplementary material.
We note that we are matching the gradient of the edges instead of the signal, though these are closely connected. 
%We also point out that terms added to the desired gradient matching are optimized and may dominate the result. 
While there is potential for a closer approximation, our method is sufficient for most use cases. %These empirical results are presented next.

%% file: chapters/experiments.tex
\section{Experiments}
\label{sec:experiments}

\begin{table}[tb]
    \centering
    \begin{tabular}{c c c c c c c}
         \toprule
         Dataset & Graphs & Nodes (avg.) & Edges (avg.) & Features & Classes & Parameters T$\to$S\\
         \midrule
         Citeseer & $1$ & $3,327$ & $9,104$ & $3,703$ & $6$& $9.7$M$\to60$k \\
         %PubMed & $1$ & $19,717$ & $88,648$ & $500$ & $3$ \\
         Photo & $1$ & $7,650$ & $238,162$ & $745$ & $8$ & $3.7$M$\to12$k\\
         Computers & $1$ & $13,752$ & $491,722$ & $767$ & $10$ & $1.3$M$\to13$k \\
         WikiCS & $1$ & $11,701$ & $216,123$ & $300$ & $10$ & $872$k$\to5.3$k\\
         %CS & $1$ & $18,333$ & $163,788$ & $6,805$ & $15$ \\
         Physics & $1$ & $34,493$ & $495,924$ & $8,415$ & $5$ & $4.5$M$\to135$k \\
         PPI & $20$ & $2,245.3$ & $61,318.4$ & $50$ & $121$ & $3.1$M$\to39$k\\
         \bottomrule
    \end{tabular}
    \caption{Statistics of our considered benchmark datasets for node classification. }
    \label{tab:graphs}
\end{table}
We now empirically evaluate our claims using our proposed metrics and the effect of DropDistillation on Knowledge Distillation.
We evaluate our method on the six standard benchmark datasets for node classification. Details about these datasets are given in Table~\ref{tab:graphs}.
Citeseer~\citep{10.1145/276675.276685}, Photo~\citep{shchur2018pitfalls}, WikiCS~\citep{mernyei2020wiki}, Computers~\citep{shchur2018pitfalls}, Physics~\citep{shchur2018pitfalls} are transductive node classification tasks, so we randomly split the nodes into train, validation, and test sets. The same train nodes are used to optimize the target and distillation loss. The best-performing model based on the validation accuracy is chosen, and metrics are reported based on the corresponding test nodes. 
PPI~\citep{zitnik2017predicting} is an inductive multi-class classification task, so we use the public split based on entire graphs, replace the cross-entropy with the binary-cross-entropy loss and report the F1-score based on the test graphs.
Each experiment runs for five random parameter initialization, and each metric's average and standard deviation is reported.
We optimize our models using Adam~\citep{kingma2014adam} using a learning rate of $0.005$ and perform early stopping when the validation score does not improve for at least $400$ steps.
Our implementation reuses the general training framework and the existing methods from~\citep{joshi2022representation} based on Pytorch-Geometric~\citep{Fey/Lenssen/2019}\footnote{Our implementation is available at \url{https://github.com/roth-andreas/distilling-influences}.}.
We use a $3$-layer Graph Attention Network (GAT)~\citep{velivckovic2017graph} with residual connections as our base model for all presented experiments. We also provide results for all experiments using the Graph Convolutional Network (GCN)~\citep{kipf2016semi} as supplementary material.

\subsection{Empirical Validation of our Axioms}

\begin{table}[tb]
\centering
\caption{Mean and standard deviation of our proposed metrics over five runs with random parameter initializations.}
\begin{tabular}{c c c c c c}
\toprule
    Dataset & Acc./F1-score (\%) & C (\%) & ID (\%) & corr($\mathbf{id},\mathbf{s}$) & corr($\mathbf{h},\mathbf{s}$)\\  
     \midrule
    Citeseer & $57.8\pm 2.5$ & $32.2\pm3.9$ & $48.8\pm2.8$ & $0.01\pm0.02$ & $-0.06\pm0.05$\\ 
    %Pubmed & $75.9\pm 1.0$ & $5.7\pm0.8$ & $28.1\pm8.3$ & $5.7\pm0.8$ \\  
    Photo & $79.0\pm3.6$ & $26.3\pm3.3$ & $98.8\pm3.2$& $0.03\pm0.02$ & $-0.26\pm0.05$\\
    WikiCS & $77.2\pm0.8$ & $19.3\pm0.7$ & $75.7\pm4.9$& $-0.02\pm0.01$ & $-0.22\pm0.02$\\
    Computers & $65.2\pm5.5$ & $46.3\pm3.7$ & $104.9\pm4.3$& $0.05\pm0.03$ & $-0.17\pm0.03$\\ 
    %CS & $72.9\pm3.3$ & $40.4\pm4.2$ & $84.2\pm4.6$ & $61.7\pm2.6$ \\
    Physics & $86.9\pm1.8$ & $13.3\pm1.4$ & $75.5\pm7.3$& $-0.03\pm0.02$ & $-0.28\pm0.01$\\
    PPI & $84.3\pm0.2$ & $10.5\pm0.1$ & $76.6\pm2.1$ & - & -\\
     \bottomrule
\end{tabular}
\label{tab:inf}
\end{table}

We now use our proposed metrics to empirically validate the presented axioms without considering the application of knowledge distillation.
All one-hop neighbors are used as context nodes $C_i=N_i$ for our metrics.
We provide the results for accuracy or F1-score, churn $C$, influence difference $ID$, the correlation $\mathrm{corr}(\mathbf{id},\mathbf{s})$, and the correlation $\mathrm{corr}(\mathbf{h},\mathbf{s})$ based on pairwise models with the same hyperparameters in Table~\ref{tab:inf}. 
% Results description
While the accuracy is rather stable between runs, $ID$ demonstrates substantial differences in the influence between model instantiations of at least $48.8\%$ in relative change for an average node pair. This already shows that each model bases their predictions on different features in the data, confirming Axiom~\ref{hyp:diff}. Notably, the influence difference is still significant even when the prediction churn is relatively low.
Further, these influence differences show no correlation to the stability of node predictions. This indicates that all nodes are similarly unstable, not just those predicted differently by two models, supporting Axiom~\ref{hyp:unconditional}.
The correlation to the entropy of neighboring labels is much higher and always negative, indicating that less variance in the labels of neighboring nodes helps the stability of a prediction. This supports Axiom~\ref{hyp:redundant}. We note that this correlation is small in some cases, showing that the entropy of neighboring labels is insufficient to capture the reason for the stability fully.

These findings show that GNNs use different reasonings to make their predictions.
This aligns well with theoretical ideas on model similarity that hypothesize that each model learns different subsets of features from the data~\citep{allen2020towards}. 
It also validates our presented method, as matching the influence difference provides a meaningful metric for aligning a student to its teacher and finds signals even in the correctly classified nodes.
%This is undesirable not only for performance reasons but also for the reliability and explainability of GNNs.

\subsection{Evaluating of DropDistillation for Knowledge Distillation}

\begin{table}[tb]
\footnotesize
    \centering
    \begin{tabular}{c c c c c c c}
    \toprule
    \textbf{Accuracy/F1-score} & Computers & Physics & WikiCS & Photo & Citeseer & PPI\\
         %Model & Acc(\%) & Acc(\%) & Acc(\%) & Acc(\%) & Acc(\%) & F1(\%)\\
         \midrule
         Teacher & $80.8$ & $91.2$ & $79.7$ & $85.4$ & $68.8$ & $98.8$ \\
         \midrule
         Student & $65.2\pm5.5$ &
         $86.9\pm1.8$ & 
         $77.2\pm0.8$ & 
         $79.0\pm3.6$ & 
         $57.8\pm2.5$ & 
         $84.5\pm0.3$
         \\
         Student+DropEdge & $72.2\pm2.5$ &
         $88.7\pm1.3$ & 
         $77.9\pm0.5$ & 
         $82.4\pm3.3$ & 
         $60.9\pm2.7$ & 
         $82.4\pm0.2$
         \\
         KD & $67.5\pm5.6$ &
         $87.1\pm1.9$ & 
         $78.8\pm0.7$ &
         $78.6\pm3.6$ &
         $59.8\pm2.1$ &
         $84.7\pm0.3$
         \\
         KD+DropEdge & $\underline{77.5}\pm3.4$ &
         $\underline{89.4}\pm0.7$ & 
         $\underline{79.2}\pm0.6$ &
         $82.4\pm2.9$ &
         $\underline{62.9}\pm3.3$ &
         $79.0\pm0.2$
         \\
         G-CRD & $70.6\pm3.6$ &
         $87.3\pm1.2$ &
         $77.5\pm0.3$ &
         $80.8\pm1.6$ &
         $58.0\pm2.9$ &
         $\underline{84.7}\pm0.2$
         \\
         G-CRD+DropEdge & $73.7\pm2.7$ &
         $88.5\pm1.1$ &
         $78.9\pm0.2$ &
         $\underline{84.7}\pm3.5$ &
         $59.4\pm3.9$ &
         $80.9\pm0.2$
         \\
         \midrule
         DropDistillation & $\mathbf{80.0}\pm0.9$ &
         $\mathbf{90.8}\pm0.9$ &
         $\mathbf{79.6}\pm0.4$ &
         $\mathbf{84.9}\pm2.7$ &
         $\mathbf{66.3}\pm0.9$ &
         $\mathbf{85.0}\pm0.1$
         \\
         \bottomrule
    \end{tabular}
    \caption{A comparison of the performance on the node classification tasks. For PPI, the F1-score is reported, and in all other cases, accuracy is reported. The best results are indicated in bold, the second-best are underlined. }
    \label{tab:kd_acc}
\end{table}

We now evaluate our proposed DropDistillation (DD) on several benchmark tasks for Knowledge Distillation in node classification. We train one teacher model and five student models for each constellation based on the same teacher. We report the average prediction churn of a student compared to the teacher and the task performance. 
In our experiments, both the teachers and the students are GATs. To demonstrate the versatility of DropDistillation, we additionally present results for all experiments using a GCN as the student model in the supplementary material. 

%Between the teacher and the student, we change 
The GATs for teacher and student models only differ in the multiplier $q$ on the number of hidden channels per layer and the number of attention heads. 
For the teacher model, we select the best-performing $q$ and the number of heads fitting into our GPU memory of $8$ GB. For the student model, we choose a much smaller $q$ and number of heads with a noticeable difference in performance, allowing us to evaluate the impact of different methods. To only evaluate the effect of the distillation methods, we keep these fixed across all methods. Table~\ref{tab:graphs} reports the number of parameters for all models. Compression factors are at least $40\times$ between teachers and students.

\subsubsection{Comparable Methods and Hyperparameters}
We evaluate several state-of-the-art student models as proposed originally as baselines and directly compare each model with its regularized version. We use grid search with the parameters described for each method below.
The following methods are considered:
\begin{itemize}
    \item As a baseline, we optimize a plain Student (\textbf{Student}) only having access to the true labels and no signal from the teacher model.
    \item The current state-of-the-art in terms of minimizing prediction churn directly performs Knowledge Distillation (\textbf{KD})~\citep{hinton2015distilling}, as this provably optimizes churn~\citet{jiangchurn}. We tune the parameter $\alpha \in \{0.25,0.5\}$.
    \item We also compare our method to Graph Contrastive Representation Distillation (\textbf{G-CRD})~\citep{joshi2022representation}, which we consider to be the current state-of-the-art in terms of distillation regarding downstream performance. G-CRD matches each node's student representation to its representation in the teacher model while separating representations of different nodes. We tune its parameter $\beta \in \{0.03,0.1,0.3\}$.
\end{itemize}
To quantify the advantage of DropDistillation, we additionally present results when combining each of these three methods with DropEdge. We consider the dropout rates in $\{0.2,0.4\}$ and apply DropEdge before each convolution of the student during training. 

For DropDistillation, we combine it with KD and additionally only tune the number of iterations to use $\mathcal{L}_{DD}$ in $\{50,800,1500\}$ and use the same dropout rate $p^*=0.2$ across all experiments, as we find DD to be sufficiently stable. 
As the goals of DropEdge are complementary to ours, we also include it in our grid search. 
%As we do not want to clutter our result tables, we report the average churn of five students compared to the same teacher for immediate interpretability.
%We report the mean and standard deviation for all metrics based on five random instantiations of student models.

\begin{table}[tb]
\footnotesize
    \centering
    \begin{tabular}{c c c c c c c}
    \toprule
    \textbf{Churn} $\mathbf{C}$ & Computers & Physics & WikiCS & Photo & Citeseer & PPI\\
         \midrule
         Student & $38.3\pm5.5$ &
         $10.3\pm1.3$ &
         $16.9\pm0.5$ & 
         $22.3\pm2.8$ &
         $29.3\pm3.1$ &
         $9.0\pm0.2$
         \\
         Student+DropEdge & $29.7\pm2.7$ &
         $8.5\pm0.8$ &
         $16.3\pm0.6$ & 
         $18.9\pm2.7$ &
         $27.6\pm2.9$ &
         $10.3\pm0.1$
         \\
         KD & $32.1\pm4.8$ & 
         $9.0\pm2.1$ &
         $9.3\pm0.4$ &
         $22.9\pm3.3$ &
         $30.9\pm5.3$ &
         $\underline{9.0}\pm0.1$\\
         KD+DropEdge & $\underline{21.6}\pm2.5$ &
         $\underline{6.4}\pm1.0$ &
         $\underline{9.2}\pm0.4$ &
         $19.2\pm2.9$ &
         $\underline{24.6}\pm2.7$ &
         $12.0\pm0.1$\\
         G-CRD & $30.0\pm2.5$ &
         $11.6\pm1.9$ &
         $15.5\pm0.9$ &
         $19.8\pm2.5$ &
         $34.4\pm2.9$ &
         $9.1\pm0.1$\\
         G-CRD+DropEdge & $27.6\pm3.8$ &
         $10.0\pm1.7$ &
         $13.6\pm0.6$ &
         $\underline{16.7}\pm3.5$ &
         $33.9\pm2.5$ &
         $11.3\pm0.2$\\
         \midrule
         DropDistillation & $\mathbf{19.1}\pm1.7$ &
         $\mathbf{4.3}\pm1.0$ &
         $\mathbf{9.0}\pm0.3$ &
         $\mathbf{15.1}\pm1.7$ &
         $\mathbf{20.5}\pm1.7$ &
         $\mathbf{8.8}\pm0.1$\\
         \bottomrule
    \end{tabular}
    \caption{Average model churn $C(S,T)$ between the teacher and each student. The models are the same as in Table~\ref{tab:kd_acc}. Lower scores are better.}
    \label{tab:kd_churn}
\end{table}

\subsubsection{Results} 
%We present the results based on the best validation score for all experiments. 
We present the test accuracies and F1-scores of the best-performing models based on the validation scores in Table~\ref{tab:kd_acc}. Our method improves the performance in all considered cases by values between $0.2\%$ and $3.4\%$. 
Results for the prediction churn of the same students and their teachers are presented in Table~\ref{tab:kd_churn}. Our method achieves even more significant improvements of up to $4.1\%$. We observed the least significant difference for PPI, for which we found the student incapable of fitting the training data.
The best results for the other methods and DD are frequently achieved when combined with DropEdge, showing that DD does not interfere with existing advancements.
We also find DD to be less prone to overfitting, which we trace back to ID being similar for stable and unstable nodes (Axiom~\ref{hyp:unconditional}).
The results for students based on the GCN are provided as supplementary material, as they exhibit similar insights.

Our experiments confirm that using our proposed DD to match the influence of the predictions between a teacher and its student is a valuable addition to KD. The predictions are matched more closely in all considered cases, increasing overall performance across all experiments. This indicates that more of the actual knowledge from the teacher can be transferred to the student. Instead of having to come up with reasons for the predictions of the teacher, transferring these influences aids the learning process of the student. %Thus, our experiments suggest that future approaches to Knowledge Distillation should match not only the outputs and representations but also the influence.

%% file: chapters/conclusion.tex
\section{Conclusion}
\label{sec:conclusion}
This work investigated the reasons behind prediction churn in graph neural networks by quantifying differences in the influence on a prediction between models based on our proposed Influence Difference (ID) metric.
These instabilities are not limited to nodes with unstable predictions but are similarly observed for stable predictions across models.
Based on these valuable signals, we propose DropDistillation (DD), a fast approximation to minimize ID between a student and its teacher in Knowledge Distillation (KD). 
Our experiments confirm the importance of aligning influences, as it improves both the stability of the student's predictions and the overall performance. 

% Future Work
%Our study paves the way for several promising future research directions. 
Our work suggests that future methods on KD can greatly benefit from incorporating similar influence-matching strategies.
%We suspect the behavior of models for other data types to be similar, which should be confirmed by similar investigations, e.g., for images and language.
Beyond the scope of KD, we see great potential in enhancing our understanding of how models can effectively leverage the diverse features inherent in the data. 
%Constructing models that capture all views in the data can achieve better performance and robustness. 

%% file: chapters/appendix.tex
\section{Mathematical Details}
\label{sec:appendix}
In this part, we provide the proofs for the Propositions in the main paper.

\subsection{Proof of Proposition 1.}
Let $G=(V,E)$ be a graph with each node $v_i$ having two neighbors, denoted $v_i^1$ and $v_i^2$. Further, let all node features be initialized such that both neighbors start with the same state $x_i^1 = x_i^2$. 
In $f$, all nodes $x_i^1$ are weighted by $p$, and all nodes $x_i^2$ are weighted by $\epsilon$. In $g$, weights are exchanged. Thus, results for $f$, $g$ are the same and $C(f,g)=0$. However, the influence difference is large, precisely for $p>3\epsilon$, it is
\begin{equation}
    \mathrm{ID}(f,g) = \frac{p-\epsilon}{0.25\cdot (p+\epsilon)} > 2\, .
\end{equation}
\null\nobreak\hfill\ensuremath{\square}

\subsection{Proof of Proposition 2.} Our proof closely follows the proof for random noise by Srinivas et al. We use the Taylor-approximation of T and S around the point $(\mathbf{X},\mathbf{A})$ and use our assumption about zero mean for each entry of $\mathbf{A}_{\mathrm{drop}}$.

\begin{align*}
    & \mathbb{E}_{\mathbf{A}_\mathrm{drop}}\left[\sum_{v,i=1}^{N,C}(T(\mathbf{X}, \mathbf{A}+\mathbf{A}_{\mathrm{drop}})_{vi} - S(\mathbf{X},\mathbf{A}+\mathbf{A}_{\mathrm{drop}})_{vi})^2\right] \\
    &= \mathbb{E}_{\mathbf{A}_\mathrm{drop}}\left[\sum_{v,i=1}^{N,C}(T(\mathbf{X}, \mathbf{A})_{vi} +vec(\nabla_\mathbf{A}T(\mathbf{X},\mathbf{A}))^Tvec(\mathbf{A}_{\mathrm{drop}})) + \mathcal{O}(vec(\mathbf{A}_{\mathrm{drop}}))\odot vec(\mathbf{A}_{\mathrm{drop}})))\right. \\
    &- \left. S(\mathbf{X}, \mathbf{A})_{vi} +vec(\nabla_\mathbf{A}S(\mathbf{X},\mathbf{A}))^Tvec(\mathbf{A}_{\mathrm{drop}})) + \mathcal{O}(vec(\mathbf{A}_{\mathrm{drop}}))\odot vec(\mathbf{A}_{\mathrm{drop}}))))^2\vphantom{\sum_1^2}\right] \\
    &= \mathbb{E}_{\mathbf{A}_\mathrm{drop}}\left[\sum_{v,i=1}^{N,C}\left(T(\mathbf{X}, \mathbf{A})_{vi} - S(\mathbf{X}, \mathbf{A})_{vi}\right)^2\right] \\
    &+ \mathbb{E}_{\mathbf{A}_\mathrm{drop}}\left[(vec(\nabla_\mathbf{A}T(\mathbf{X},\mathbf{A})_{vi})^Tvec(\mathbf{A}_{\mathrm{drop}})) - vec(\nabla_\mathbf{A}S(\mathbf{X},\mathbf{A}))_{vi})^Tvec(\mathbf{A}_{\mathrm{drop}})))^2\right] \\
    &+ \mathbb{E}_{\mathbf{A}_\mathrm{drop}}\left[\sum_{v,u=1}^{N,N}\mathcal{O}((\mathbf{A}_{\mathrm{drop}})_{vu}^2)\right] \\
    &= \sum_{v,i=1}^{N,C}(T(\mathbf{X})_{vi} - S(\mathbf{X})_{vi})^2 \\
    &+ \mathbb{E}_{\mathbf{A}_\mathrm{drop}}\left[ \left(vec(\nabla_{\mathbf{A}} T(\mathbf{X},\mathbf{A})_{vi} - \nabla_{\mathbf{A}} S(\mathbf{X},\mathbf{A})_{vi})^T vec(\mathbf{A}_{\mathrm{drop}})\right)^2 + \sum_{v,u=1}^{N,N}\mathcal{O}((\mathbf{A}_{\mathrm{drop}})_{vu}^2)\right]
\end{align*}

All terms linear in $\mathbf{A}_{\mathrm{drop}}$ have expectation zero, as $\mathbb{E}\left[(\mathbf{A}_{\mathrm{drop}})_{uv}\right] = 0$ for all $u,v\in [1,\dots,N]$.

\section{Additional Experiments using the GCN}
\begin{table}[tb]
\centering
\caption{Mean and standard deviation of our proposed metrics over five runs with random parameter initializations.}
\begin{tabular}{c c c c c c}
\toprule
    Dataset & Acc./F1-score (\%) & C (\%) & ID (\%) & corr($\mathbf{id},\mathbf{s}$) & corr($\mathbf{h},\mathbf{s}$)\\  
     \midrule
    Citeseer & $54.3\pm 2.7$ & $32.2\pm3.9$ & $47.8\pm4.9$ & $-0.03\pm0.07$ & $-0.08\pm0.03$\\ 
    %Pubmed & $75.9\pm 1.0$ & $5.7\pm0.8$ & $28.1\pm8.3$ & $5.7\pm0.8$ \\  
    Photo & $56.5\pm12.2$ & $59.0\pm8.5$ & $55.6\pm11.5$& $0.00\pm0.06$ & $-0.08\pm0.06$\\
    WikiCS & $71.1\pm1.0$ & $29.7\pm2.6$ & $21.7\pm5.7$& $-0.06\pm0.02$ & $-0.12\pm0.01$\\
    Computers & $47.0\pm13.1$ & $71.4\pm7.2$ & $64.6\pm23.2$& $0.00\pm0.09$ & $-0.03\pm0.03$\\ 
    %CS & $72.9\pm3.3$ & $40.4\pm4.2$ & $84.2\pm4.6$ & $61.7\pm2.6$ \\
    Physics & $90.0\pm2.1$ & $13.6\pm3.9$ & $29.1\pm10.4$& $-0.01\pm0.08$ & $-0.22\pm0.06$\\
    PPI & $72.0\pm0.1$ & $10.7\pm0.1$ & $19.2\pm0.5$ & - & -\\
     \bottomrule
\end{tabular}
\label{tab:inf_gcn}
\end{table}
In this section, we provide an evaluation of the experiments shown in the main paper replacing the GAT layers with GCN layers.
The experimental setup remains the same with the teacher being the same high-capacity GAT model. The motivation for employing a simpler student model stems from it being computationally more memory and runtime efficient during inference. However, due its inferior expressivity, the GCN may not be able to match influence.  
The extent of influence differences is generally unclear as the GCN uses fixed edge weights only based on the node degrees. %expect generally less influence difference as all neighbors are treated the exact the same.

The results for our metrics for GCN models are shown in Table~\ref{tab:inf_gcn}. The influence difference is still very noticeable across all datasets, albeit less pronounced. It is again not correlated to the stability of a node prediction. The correlation to the entropy of the neighboring node labels is larger in all cases, though it is rather weak.

The effects on Knowledge Distillation are presented in Table~\ref{tab:kd_acc_gcn}. Here, the results demonstrate a higher degree of variance. For the Photo dataset, accuracy is improved by $16.0\%$ and for the Computers dataset by $14.6\%$. 
These results indicate a large potential in guiding less expressive models toward desired solutions.
However, DD is not always as effective as the accuracy is slightly behind the best other method for three datasets.
Results regarding prediction churn are presented in Table~\ref{tab:kd_churn_gcn}. Again, DD achieves large improvements for some datasets but is ineffective for others.
A similar influence may not lead to optimal results for models with different expressive power.

\begin{table}[tb]
\footnotesize
    \centering
    \begin{tabular}{c c c c c c c}
    \toprule
    \textbf{Accuracy/F1-score} & Computers & Physics & WikiCS & Photo & Citeseer & PPI\\
         %Model & Acc(\%) & Acc(\%) & Acc(\%) & Acc(\%) & Acc(\%) & F1(\%)\\
         \midrule
         Teacher & $80.8$ & $91.2$ & $79.7$ & $85.4$ & $68.8$ & $98.8$ \\
         \midrule
         Student & $47.0\pm13.1$ &
         $90.0\pm2.1$ & 
         $71.1\pm1.0$ & 
         $56.5\pm12.2$ & 
         $54.3\pm2.7$ & 
         $\underline{72.0}\pm0.1$
         \\
         Student+DropEdge & $48.0\pm12.6$ &
         $\underline{90.8}\pm1.1$ & 
         $71.6\pm1.1$ & 
         $58.8\pm10.5$ & 
         $\mathbf{57.5}\pm2.7$ & 
         $70.4\pm0.1$
         \\
         KD & $47.6\pm13.8$ &
         $90.2\pm2.0$ & 
         $73.5\pm0.8$ &
         $61.2\pm7.9$ &
         $\underline{57.0}\pm2.1$ &
         $\mathbf{72.1}\pm0.3$
         \\
         KD+DropEdge & $46.0\pm9.2$ &
         $\mathbf{91.0}\pm1.4$ & 
         $73.7\pm0.5$ &
         $60.3\pm8.2$ &
         $56.6\pm1.6$ &
         $70.5\pm0.2$
         \\
         G-CRD & $\underline{49.9}\pm12.4$ &
         $89.6\pm1.9$ &
         $73.5\pm0.6$ &
         $62.1\pm13.5$ &
         $54.3\pm2.4$ &
         $70.9\pm0.2$
         \\
         G-CRD+DropEdge & $49.1\pm10.4$ &
         $90.8\pm1.4$ &
         $\underline{73.8}\pm0.6$ &
         $\underline{62.5}\pm14.3$ &
         $56.2\pm2.2$ &
         $68.0\pm0.4$
         \\
         \midrule
         DropDistillation & $\mathbf{63.7}\pm4.5$ &
         $88.0\pm1.2$ &
         $\mathbf{74.3}\pm0.7$ &
         $\mathbf{78.5}\pm6.5$ &
         $\underline{57.0}\pm2.8$ &
         $71.9\pm0.2$
         \\
         \bottomrule
    \end{tabular}
    \caption{ comparison of the performance on the node classification tasks. For PPI, the
F1-score is reported, and in all other cases, accuracy is reported. The best results
are indicated in bold, the second-best are underlined. }
    \label{tab:kd_acc_gcn}
\end{table}

\begin{table}[tb]
\footnotesize
    \centering
    \begin{tabular}{c c c c c c c}
    \toprule
    \textbf{Churn} $\mathbf{C}$ & Computers & Physics & WikiCS & Photo & Citeseer & PPI\\
         \midrule
         Student & $57.0\pm14.2$ &
         $10.8\pm3.3$ &
         $26.2\pm1.4$ & 
         $44.8\pm10.9$ &
         $41.5\pm1.8$ &
         $\underline{15.7}\pm0.1$
         \\
         Student+DropEdge & $59.6\pm12.9$ &
         $\underline{9.7}\pm3.3$ &
         $24.8\pm0.5$ & 
         $42.9\pm10.3$ &
         $\mathbf{37.2}\pm3.6$ &
         $16.3\pm0.1$
         \\
         KD & $57.7\pm12.2$ & 
         $12.9\pm5.3$ &
         $19.6\pm0.4$ &
         $41.5\pm7.7$ &
         $60.1\pm3.2$ &
         $\underline{15.6}\pm0.1$\\
         KD+DropEdge & $59.8\pm10.4$ &
         $11.8\pm2.7$ &
         $\underline{19.0}\pm0.5$ &
         $41.3\pm7.7$ &
         $\underline{57.3}\pm8.0$ &
         $16.3\pm0.1$\\
         G-CRD & $\underline{54.8}\pm11.6$ &
         $10.8\pm1.4$ &
         $22.1\pm0.6$ &
         $\underline{40.6}\pm11.5$ &
         $44.3\pm2.2$ &
         $16.3\pm0.1$\\
         G-CRD+DropEdge & $55.4\pm7.7$ &
         $\underline{9.8}\pm1.7$ &
         $22.1\pm0.9$ &
         $41.0\pm14.2$ &
         $\mathbf{37.1}\pm4.7$ &
         $17.8\pm0.2$\\
         \midrule
         DropDistillation & $\mathbf{39.4}\pm9.1$ &
         $15.1\pm3.8$ &
         $\mathbf{18.4}\pm0.5$ &
         $\mathbf{22.4}\pm5.6$ &
         $51.7\pm2.3$ &
         $15.8\pm0.1$\\
         \bottomrule
    \end{tabular}
    \caption{Average model churn C(S, T ) between the teacher and each student. The models
are the same as in Table~\ref{tab:kd_acc_gcn}. Lower scores are better.}
    \label{tab:kd_churn_gcn}
\end{table}